\documentclass[10pt,letterpaper]{article}

\usepackage[letterpaper,top=0.68in,bottom=0.70in,left=0.74in,right=0.74in]{geometry}
\usepackage[T1]{fontenc}
\usepackage{amsmath,mathtools,amsthm}
\usepackage{newtxtext,newtxmath}
\usepackage{microtype}
\usepackage{graphicx}
\usepackage{xcolor}
\usepackage{booktabs}
\usepackage{tabularx}
\usepackage{array}
\usepackage{enumitem}
\usepackage[font=small,labelfont=bf,labelsep=period]{caption}
\usepackage[numbers,sort&compress]{natbib}
\usepackage{fancyhdr}
\usepackage{titlesec}
\usepackage{balance}
\usepackage{multicol}
\usepackage{stfloats}
\usepackage{url}
\usepackage[hidelinks]{hyperref}

\definecolor{MLETInk}{HTML}{171717}
\definecolor{MLETMuted}{HTML}{5B5B5B}
\definecolor{MLETRule}{HTML}{999999}
\definecolor{MLETBlue}{HTML}{6688A4}
\definecolor{MLETTeal}{HTML}{6C9D9A}
\definecolor{MLETRed}{HTML}{B66A5B}
\definecolor{MLETGold}{HTML}{D1992D}
\definecolor{MLETPaleBlue}{HTML}{EEF3F6}
\definecolor{MLETPaleRed}{HTML}{F7EFED}

\hypersetup{
  colorlinks=true,
  linkcolor=MLETInk,
  citecolor=MLETBlue,
  urlcolor=MLETBlue,
  pdftitle={When Should a Satellite Estimate Be Changed? Stress-Testing Neural Corrections for Evapotranspiration},
  pdfauthor={Marco Trotta}
}


\graphicspath{{figures/}{assets/}{../assets/}}

\newcommand{\mlet}{\textsc{MLET}}

\newcommand{\sourcepath}[1]{\path{#1}}

\setlist{nosep,leftmargin=1.35em}
\setlist[enumerate]{label=\arabic*.}
\titleformat{\section}
  {\large\bfseries\color{MLETInk}}{\thesection}{0.55em}{}
\titleformat{\subsection}
  {\normalsize\bfseries\color{MLETInk}}{\thesubsection}{0.50em}{}
\titleformat{\subsubsection}
  {\normalsize\itshape\color{MLETInk}}{\thesubsubsection}{0.45em}{}
\titlespacing*{\section}{0pt}{1.15ex plus 0.3ex}{0.55ex}
\titlespacing*{\subsection}{0pt}{0.95ex plus 0.25ex}{0.35ex}
\titlespacing*{\subsubsection}{0pt}{0.75ex plus 0.2ex}{0.25ex}

\renewcommand{\headrulewidth}{0.35pt}
\renewcommand{\headrule}{\hbox to\headwidth{\color{MLETRule}\leaders\hrule height \headrulewidth\hfill}}
\fancypagestyle{firstpage}{
  \fancyhf{}
  \fancyfoot[C]{\sffamily\small\thepage}
  \renewcommand{\headrulewidth}{0pt}
}
\fancypagestyle{plain}{
  \fancyhf{}
  \fancyhead[L]{\sffamily\scriptsize\color{MLETMuted}\mlet{} $\mid$ manuscript preprint}
  \fancyhead[R]{\sffamily\scriptsize\color{MLETMuted}Neural Corrections for Satellite ET}
  \fancyfoot[C]{\sffamily\small\thepage}
  \renewcommand{\headrulewidth}{0.35pt}
  \renewcommand{\headrule}{\hbox to\headwidth{\color{MLETRule}\leaders\hrule height \headrulewidth\hfill}}
}
\makeatletter
\let\ps@plain\ps@fancy
\makeatother

\makeatletter
\let\tableinput\@@input
\makeatother
\newcommand{\unit}{\mathrm{mm\,day^{-1}}}
\newcommand{\E}{\mathbb{E}}

\begin{document}
\onecolumn\thispagestyle{firstpage}
\noindent\color{MLETInk}\rule{\textwidth}{0.8pt}
\vspace{0.26em}
\begin{center}
{\fontsize{19}{22}\selectfont\bfseries
When Should a Satellite Estimate Be Changed?\\[-0.1em]
Stress-Testing Neural Corrections for Evapotranspiration\par}
\vspace{0.42em}\rule{\textwidth}{0.55pt}\vspace{0.58em}

{\large Marco Trotta}\\[0.46em]
\includegraphics[height=0.32in]{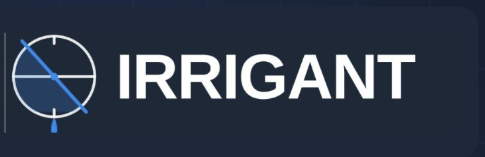}\\[0.26em]
{\small Irrigant, Idaho, USA}\\[0.12em]
{\small\href{mailto:m@irrigant.xyz}{m@irrigant.xyz}}\\[0.35em]
{\small September 25, 2026}
\end{center}
\vspace{0.35em}
\begin{center}{\large\bfseries Abstract}\end{center}
\begin{quote}\small
Neural residuals can improve satellite evapotranspiration (ET) estimates, but selectors must predict when a correction helps and reject unsupported inputs.
We evaluate ten-member models on 16,366 flux-tower observations from 151 stations paired with OpenET, across nine rolling years and five spatial folds.
At one held-out station, Gain accepted corrections on all 32 physically invalid records: it predicted a mean benefit of 0.83 $\unit$, but the corrections increased mean absolute error by 21.6 $\unit$ versus OpenET.
On spatially held-out unit errors, SupportGain reduced station-macro MAE versus Gain by 0.148 $\unit$ under wind x3.6 (simultaneous 95\% interval, 0.070 to 0.226), with 9.3\% acceptance versus Gain's 51.8\%; on clean inputs, its 0.006 $\unit$ advantage had an interval that includes zero.
These fault analyses are exploratory; none of 40 preplanned temporal comparisons passed Holm correction, while a separate predeclared cropland contrast found 0.041 $\unit$ lower station-macro MAE with crop-only training (95\% interval, 0.009 to 0.079).
\end{quote}

\section{Introduction}
OpenET already estimates actual evapotranspiration (ET).
A neural correction can move that estimate closer to a flux-tower measurement, or farther away.
The selector must answer a sharper question than whether its neural model is uncertain: will this correction improve the estimate already available?

That decision becomes harder when weather inputs fall outside the conditions represented in training.
A selector may predict benefit from familiar patterns while missing an unsupported input.
This paper tests whether a simple support rule can catch such cases, and whether any apparent advantage holds on clean observations.

We evaluate a ten-network residual model on 16,366 screened records from 151 stations.
Rolling tests cover 2012 through 2020, but reuse stations across years; they do not test unseen-site transfer.
Every rejected correction returns the original OpenET estimate, and the score includes every test record.
This prevents rejection from appearing accurate by hiding difficult cases.

The evidence gives a mixed answer.
Clean data do not establish a reliable advantage for support screening over learned benefit prediction.
An exploratory station-level failure shows how far a confident correction can miss, while controlled weather changes reveal when support screening can reduce error by rejecting many corrections.
A predeclared cropland comparison also favors training on cropland records, subject to method-selection uncertainty.

The study makes three contributions.
First, it evaluates a correction by how much error it removes from the existing satellite estimate, using spatially cross-fitted selector data.
Second, it separates one observed input failure from controlled weather transformations and from the wider record of physical rule violations.
Third, it compares all-station and crop-only training on the same held-out cropland records.
These are retrospective prediction results; they do not identify irrigation status or measure irrigation response.
\clearpage\twocolumn
\section{Related work}
Selective regression asks when a model should defer to an available alternative.
Prior work studies multi-expert deferral and conformalized selective prediction \citep{mao2024,sokol2026}.
Here, the alternative is the existing OpenET estimate.
The selector predicts improvement over that estimate, and evaluation counts the fallback on rejected records.
The decision rule is established; our contribution is its environmental stress test, not a new deferral theory.

Environmental prediction methods use input distance and cross-fitting to assess whether a record resembles the training data \citep{meyer2021,nguyen2026}.
Ensemble disagreement can also change outside the training distribution \citep{laksh2017,mathelin2023}, while neural networks may extrapolate along unsupported inputs \citep{xu2021}.
We test a nearest-neighbor distance screen beside these learned scores.
It is a simple check, not a calibrated guarantee of reliable predictions.

Machine-learning methods already estimate and correct ET using water-balance supervision, physical covariates, and geographic transfer \citep{hascoet2022,rozanov2025,shi2025,residual2026,wei2026}.
OpenET and the associated flux archive supply the satellite estimates and measurements used here \citep{volk2024,openetdata,fluxdata}.
Because the same archive supports this evaluation, our results do not independently validate OpenET development.
Land-cover-specific OpenET biases also motivate testing a cropland training population \citep{openetissues}.

Environmental records are spatially dependent, so random row splits can overstate transfer \citep{roberts2017}.
We group nearby stations for inner selection and report a separate spatial holdout.
The rolling-year analysis tests later dates at observed stations, not new locations.

Reference ET describes a standardized surface; the target here is measured actual ET.
We use reference ET only as an input and make no claim about crop water need, irrigation response, or reference-ET forecasting \citep{kukal2026}.

\section{Data and evaluation}
\subsection{Observations and inputs}
We use the OpenET model archive, Zenodo record 10119477, and the processed flux archive, record 7636781.
The station-date join contains 16,447 labeled observations from 152 stations.
A physical screen retains 16,366 observations from 151 stations.
It removes 24 temperature violations, eight gridMET vapor-pressure-deficit (VPD) violations, 54 invalid weather values from the station archive, and three missing ET labels.
These counts overlap.
A rule violation does not diagnose an instrument fault.
The retained stations form 102 proximity groups at 10 km.

The target is energy-balance-corrected measured ET from the archive's \texttt{ET\_corr} field.
The observations are sparse satellite validation dates, not a daily time series.
We retain negative target values and do not clip predictions.
The Croplands subset contains 5,203 observations from 58 stations and 30 groups.
Land-cover labels do not identify irrigation status.

Let $y_{st}$ denote measured ET and $o_{st}$ the OpenET estimate at station $s$ and date $t$.
The model uses that estimate, reference ET, seasonal position, temperature, VPD, and wind:
\begin{equation}
\begin{aligned}
x_{st}&=[o_{st},ETo_{st},h_t,T_{st},VPD_{st},u_{st}],\\
h_t&=[\sin(2\pi d_t/365),\cos(2\pi d_t/365)].
\end{aligned}
\end{equation}
Here $d_t$ is day of year, $T$ is gridMET average temperature, $u$ is gridMET wind speed, and ETo is reference ET.
The seasonal encoding repeats every 365 days.
The inputs and satellite estimate refer to the target date, but we do not reconstruct their release times.
This is retrospective estimation, not forecasting.

\subsection{Rolling temporal evaluation}
For each test year from 2012 through 2020, train on rows from earlier years and test on that year.
The pooled test set has 7,842 rows from 101 stations and 64 groups.
Stations can occur in both training and test years.
This design tests temporal transfer, not transfer to unseen stations.
The cropland test subset has 3,234 rows from 49 stations and 24 groups.

Each training period uses three inner folds that withhold 10 km proximity groups.
The same saved row assignments support the all-station and crop-only training arms.
The crop-only arm filters each saved partition to Croplands rows.
Outer test labels never determine model fits, selector scores, or thresholds.
We save row identifiers for every fit, validation, and test partition.

\subsection{Scoring and uncertainty}
The primary score calculates mean absolute error (MAE) within each test station, then averages stations:
\begin{equation}
\operatorname{MAE}_{\mathrm{station}}(f)=
\frac{1}{S}\sum_{s=1}^{S}\frac{1}{n_s}\sum_{t=1}^{n_s}|y_{st}-f(x_{st})|.
\label{eq:macro}
\end{equation}
This gives each station equal weight, regardless of its number of observations.
It does not estimate area-weighted agricultural error.
Coverage is the station-weighted share of observations that receive a correction.
The complete-system score uses OpenET wherever the selector rejects a correction.

We report paired differences as baseline error minus candidate error.
Positive differences favor the candidate.
The primary and extended selector families contain 40 and 52 preplanned comparisons, respectively.
Both use Holm's multiple-test correction.
The cropland analysis has one predeclared contrast and a paired group bootstrap with 2,000 draws and seed 20261005.
Intervals resample proximity groups but keep fitted models and selectors fixed.
They omit retraining uncertainty.
Repeated split settings are stability checks, not independent trials.

\section{Neural correction models}\label{sec:models}
\subsection{Ten-network residual model}
The main gridMET and cropland experiments use an ensemble of ten residual networks.
Each network has two 32-unit ReLU layers and one scalar output.
Each model has 1,345 trainable parameters and seven input features.
Training uses Adam with learning rate 0.001, batch size 256, L2 parameter 0.001, and 120 iterations.
Each fit standardizes its inputs and training residuals.
The ten fixed seeds run from 20260713 through 20260722.
The model predicts the residual from OpenET to measured ET.
The selector uses the ensemble mean correction and its between-network spread.
\emph{Full} applies every correction; OpenET applies none.
Training does not use test-based stopping or architecture search.
We retain every fit warning in the run records.

\section{When should the correction be applied?}\label{sec:selective}
\subsection{Predicting benefit differs from predicting error}
Let $o(x)$ denote the available satellite estimate and $g(x)$ a fitted neural residual.
A selector $a(x)\in\{0,1\}$ returns:
\begin{equation}
 f_a(x)=o(x)+a(x)g(x).
\end{equation}
Every observation receives a prediction: the corrected estimate if accepted, or OpenET if rejected.
The correction's realized benefit is
\begin{equation}\label{eq:gain}
 D(x,y)=|y-o(x)|-|y-o(x)-g(x)|.
\end{equation}
Positive $D$ means the correction lowers absolute error.
For a fixed fitted model, the system's mean absolute error satisfies
\begin{equation}\label{eq:selective-risk}
 R(a)=R(o)-\E[a(X)D(X,Y)].
\end{equation}
The selector's target is the expected benefit given its inputs, $\mu(x)=\E[D(X,Y)\mid X=x]$.
It should apply the correction when $\mu(x)>0$.
A selector that predicts low neural error need not predict positive improvement over $o$.

Consider two observations.
For the first, OpenET and measured ET are both 0, while the correction is 0.1; the correction adds 0.1 error.
For the second, measured ET is 10, OpenET is 0, and the correction is 9; it removes 9 error.
The neural errors are 0.1 and 1, respectively.
A rule that rejects the larger neural error would keep the harmful correction and discard the helpful one.
Ensemble spread measures disagreement between fitted networks, not whether either correction improves on OpenET.

\subsection{Learning and checking the selectors}
\begin{figure*}[t]
\captionsetup{width=.96\textwidth}
\centering
\includegraphics[width=.98\textwidth]{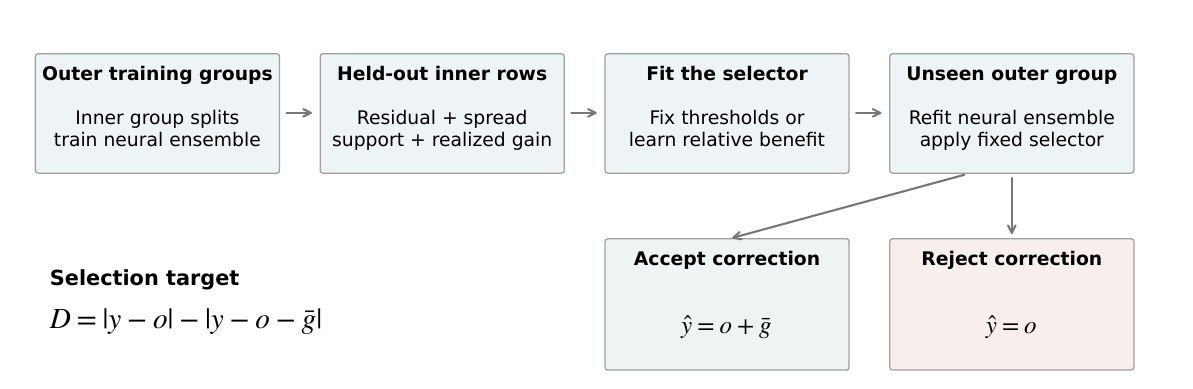}
\caption{The selectors use predictions from inner groups that the neural networks did not train on. Outer test targets measure performance only. A rejected correction returns the original OpenET estimate.}
\label{fig:selector-design}
\end{figure*}

The main experiment fixes the ten-network residual ensemble from Section~\ref{sec:models}.
Each outer training set produces three sets of neural predictions on held-out proximity groups.
These predictions train the selectors and set their thresholds.
The neural ensemble is then refitted on the full outer training set.
Outer test targets do not set model parameters, selector scores, thresholds, or input scales.
Figure~\ref{fig:selector-design} shows the flow.

For residual predictions $g_1,\ldots,g_{10}$, we use their mean $\bar g$ and sample spread:
\begin{equation}
 s_g(x)=\left[\frac{1}{9}\sum_{j=1}^{10}(g_j(x)-\bar g(x))^2\right]^{1/2}.
\end{equation}
This is a mean-only ensemble inspired by \citet{laksh2017}.
The input-support score is the average distance to the five nearest training observations:
\begin{equation}
 d_5(x)=\frac{1}{5}\sum_{z\in N_5(x)}\|S(x)-S(z)\|_2,
\end{equation}
where $S$ standardizes inputs using the training set and $N_5$ contains the five nearest training observations.
We rebuild the neighbor index within each training partition and set thresholds from held-out inner distances.
This simple screen follows prior work on training support \citep{meyer2021}, but it does not guarantee reliable predictions under shift.

\emph{Spread95} accepts observations below the station-weighted 95th percentile of inner ensemble spread.
\emph{Support95} applies the same rule to input-support distance.
\emph{Gain} predicts the benefit in Equation~\ref{eq:gain} with a fixed boosted-tree regressor.
Its six inputs are $(\bar g,|\bar g|,s_g,\log(1+d_5),o,\mathrm{ETo})$.
It uses 80 iterations, learning rate 0.05, seven leaves, minimum leaf size 50, L2 parameter 10, and no early stopping.
Training weights give equal total weight to each station and are normalized to mean one.
Gain accepts a correction only when predicted benefit is positive.
\emph{SupportGain} also requires Support95 to accept.
The neural estimator is the same for every selector.

MonoGain requires predicted benefit to decrease as ensemble spread and support distance increase.
AugmentedGain adds randomized weather faults to benefit-model training.
The extended comparison adds TunedGain, LogisticSign, and ConformalCSR under a frozen protocol.
Uniform, LocalShrinkage, and Clip appear in the earlier exploratory analysis in Appendix~\ref{app:measured-weather}.
No selector hyperparameters change after the outer results are viewed.

\subsection{Earlier measured-weather probes}
An earlier analysis compares the seven archived inputs with a six-input version that omits VPD.
It uses both proximity withholding and joint proximity-and-time withholding.
Each natural comparison scores every test observation, including OpenET fallbacks, and reports cropland results separately.
The 95\% inner threshold does not guarantee 95\% acceptance on test observations.

Additional probes multiply VPD or wind by 10, or add $20\,{}^\circ\mathrm{C}$ to temperature.
They change the test inputs while holding labels and fitted parameters fixed.
Only rows with nonnegative exported actual vapor pressure are eligible.
The VPD probe is a negative control when the model omits VPD.
These transformations are not weather interventions and do not estimate fault prevalence.
The protocol sets each magnitude before the new run.

\section{Results}\label{sec:primary-results}
\subsection{One station shows how a correction can fail}
In one exploratory three-network run, Gain accepts all 32 records with physically invalid weather at the held-out \texttt{manilacotton} station.
It predicts a mean benefit of 0.83 $\unit$.
Instead, the corrections raise station-macro MAE from OpenET's 1.095 to 22.681 $\unit$, a 21.586 $\unit$ increase.
SupportGain rejects all 32 corrections and returns OpenET.
The mean correction magnitude is 21.98 $\unit$, far above the inner 95th percentile of 1.17 $\unit$.
Ensemble spread is also above its inner 95th percentile: 39.42 versus 0.37 $\unit$.
This is one station and one group, with no multi-group interval.

We then exclude that station and compare 133 other rule-violating records from 21 stations in 17 groups.
SupportGain has 0.015 $\unit$ lower MAE than Gain (95\% group-bootstrap interval [0.000, 0.034]).
Only two groups show a difference; 15 show none.
OpenET has lower MAE than either selector: 0.755 $\unit$, compared with 0.791 for Gain and 0.776 for SupportGain.
These records do not establish a device fault or a broad selector advantage.

\begin{figure*}[t]
\captionsetup{width=.96\textwidth}
\centering\includegraphics[width=.96\textwidth]{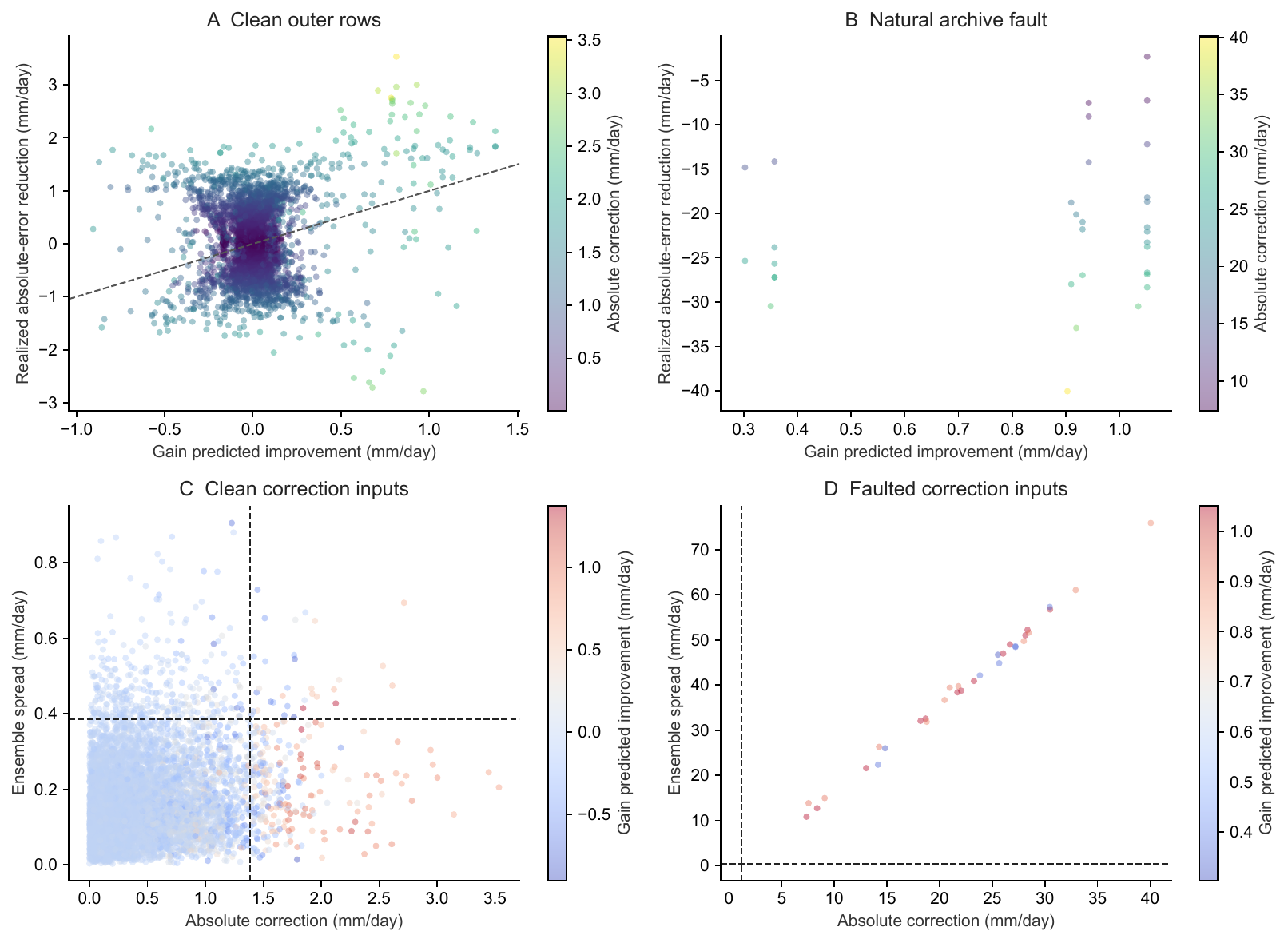}
\caption{Each point is one of 32 records from the held-out station. Gain predicts benefit even though correction size and ensemble spread exceed their inner-training 95th percentiles.}
\label{fig:gain-mechanism}
\end{figure*}

\begin{figure*}[t]
\captionsetup{width=.96\textwidth}
\centering\includegraphics[width=.96\textwidth]{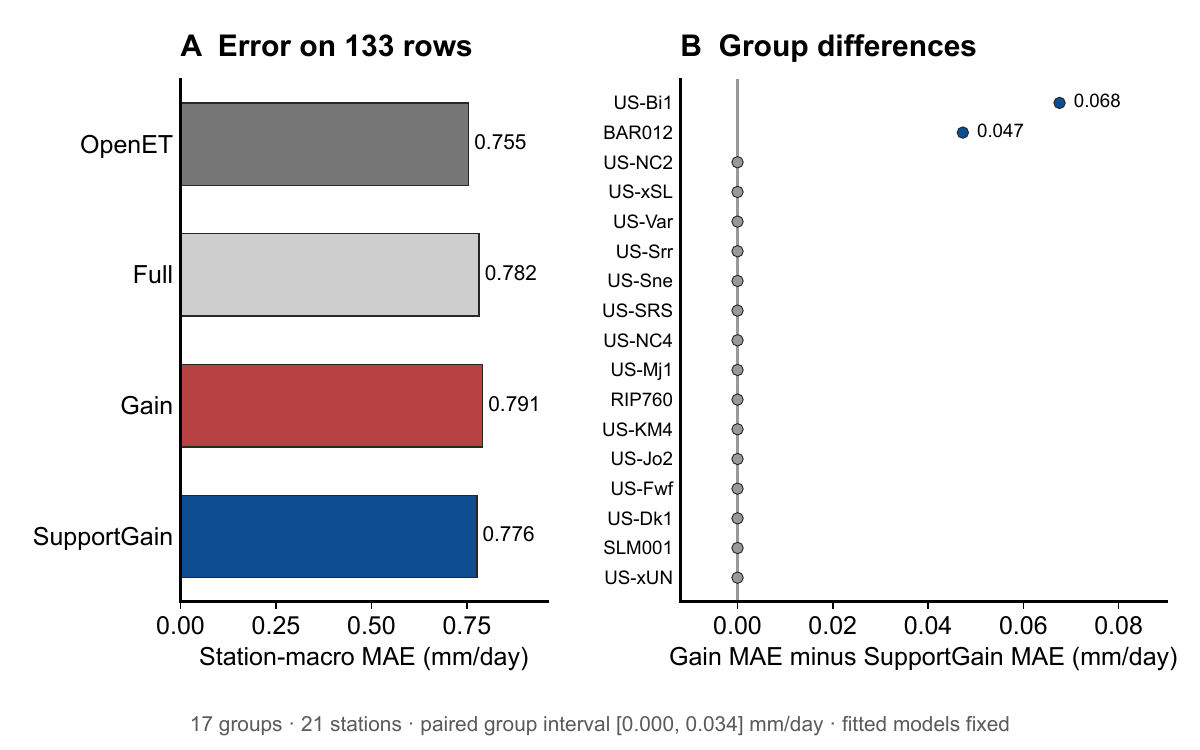}
\caption{Among 133 rule-violating records outside the known station, only two of 17 groups show a difference between Gain and SupportGain. OpenET has the lowest station-macro MAE. Rejected corrections return OpenET.}
\label{fig:natural-violations}
\end{figure*}

\subsection{Support screening helps under altered inputs, but accepts fewer records}
The all-station protocol defines 40 comparisons before fitting.
None has a Holm-adjusted one-sided $p$-value below 0.05; the smallest is 0.126.
These tests do not directly compare SupportGain with Gain.

Across 30 shared split-seed and proximity settings, SupportGain lowers clean MAE over Gain by a mean of 0.00438 $\unit$ (standard deviation 0.00289).
Only one of 30 nominal group-bootstrap intervals excludes zero.
The settings reuse observations, so they do not show a stable clean advantage.

\begin{figure*}[t]
\captionsetup{width=.96\textwidth}
\centering\includegraphics[width=.98\textwidth]{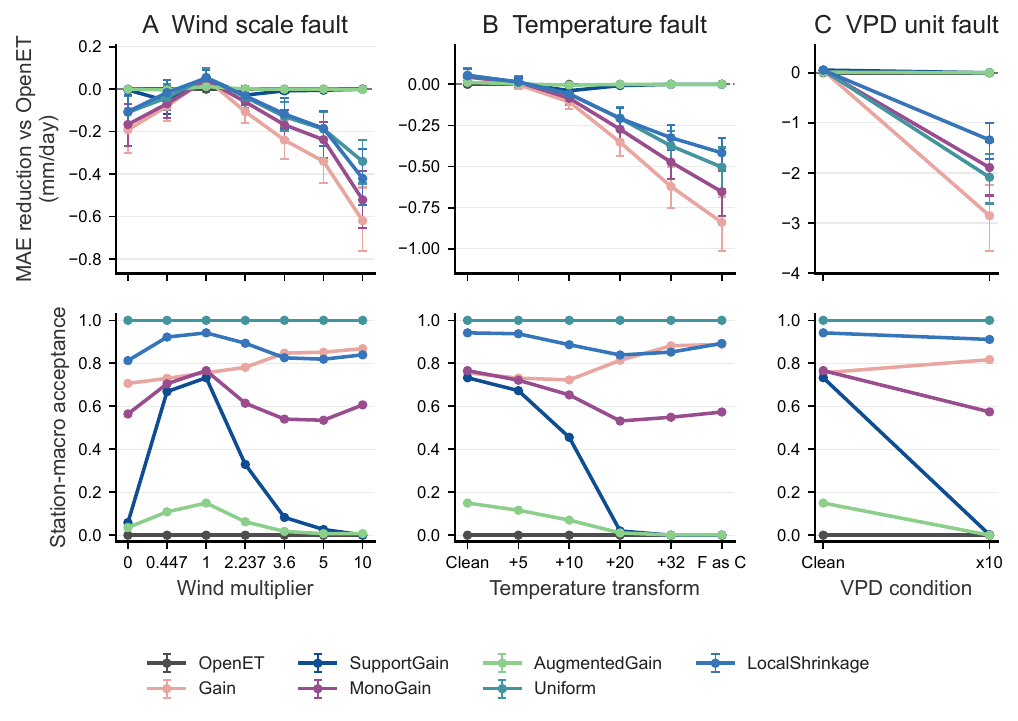}
\caption{Ten-network results under fixed weather changes. The top row shows MAE reduction over OpenET; the bottom row shows station-weighted acceptance. Descriptive 95\% intervals condition on fitted models. Wind labels give multipliers; temperature labels give Celsius offsets. ``F as C'' reads Fahrenheit values as Celsius. The VPD panel shows one kPa-to-hPa error.}
\label{fig:gridmet-weather-faults}
\end{figure*}

SupportGain has lower point MAE than Gain in all 30 settings under each of twelve synthetic weather transforms.
Across the 30 settings, temperature increased by 32 $^\circ$C yields a mean improvement of 0.55623 $\unit$ (SD 0.04753; range 0.44878 to 0.62684).
These comparisons reuse observations and keep labels and fitted models fixed.
They describe sensitivity to the chosen transforms, not natural fault prevalence.
Figure~\ref{fig:gridmet-weather-faults} shows the wind, temperature, and VPD probes.
The VPD panel tests a kPa-to-hPa unit error at one magnitude.

A separate post hoc five-fold spatial holdout excludes each test station from its fold's fit.
Across 102 groups, the clean Gain-minus-SupportGain difference is 0.0059 $\unit$ (simultaneous 95\% interval [-0.0069, 0.0186]).
Under wind multiplied by 3.6, SupportGain lowers MAE by 0.1478 $\unit$ (simultaneous 95\% interval [0.0697, 0.2260]).
It accepts 9.3\% of records, compared with 51.8\% for Gain.
Coverage falls below 0.3\% under Fahrenheit-as-Celsius and VPD multiplied by 10.
This exploratory analysis uses the same archive and conditions on fitted models; Appendix~\ref{app:spatial-holdout} reports the full estimates.

\begin{figure*}[t]
\captionsetup{width=.96\textwidth}
\centering\includegraphics[width=.96\textwidth]{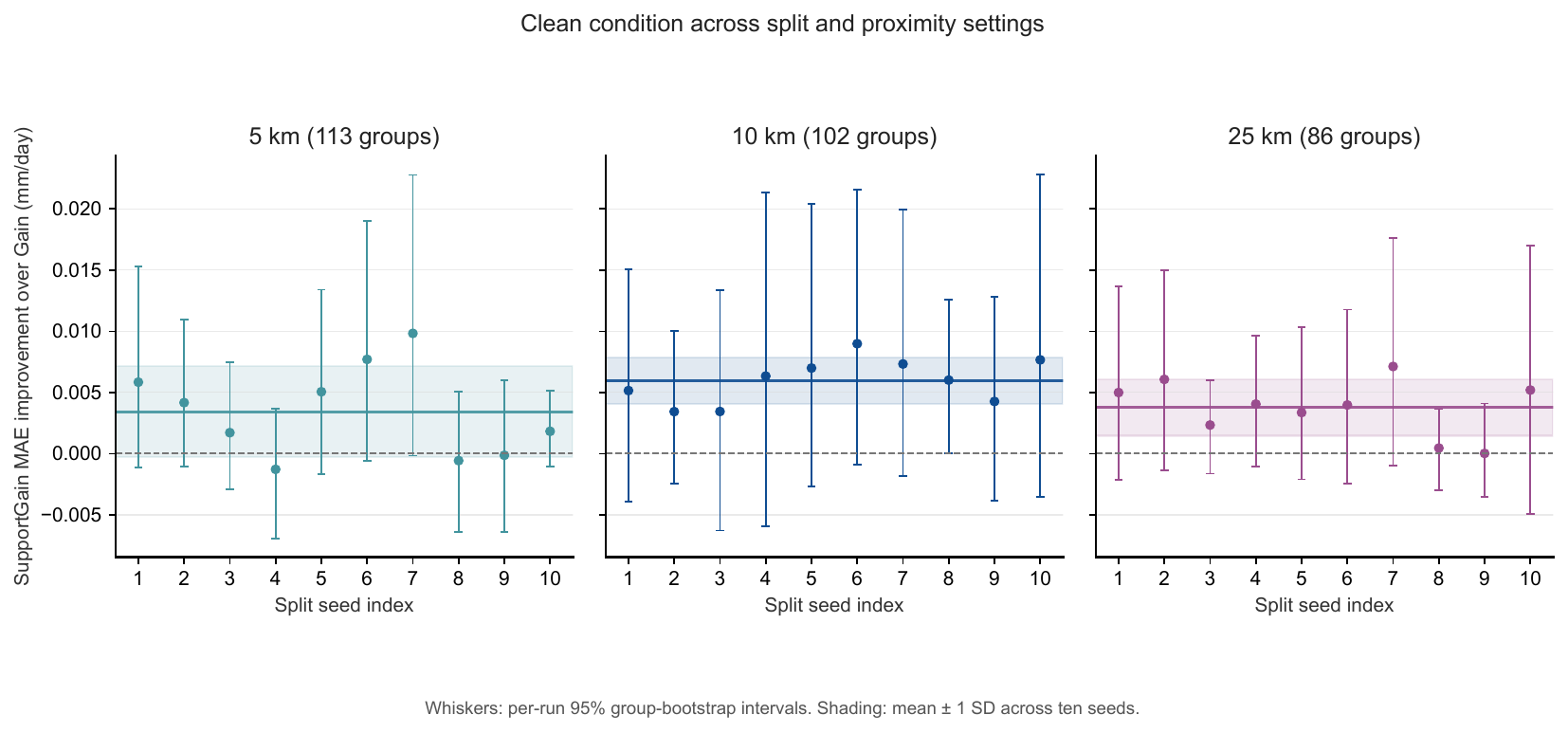}
\caption{Clean SupportGain improvement over Gain for ten inner split seeds and three proximity thresholds. Whiskers show 95\% group-bootstrap intervals; shading shows the mean plus or minus one standard deviation across split seeds.}
\label{fig:gridmet-split-sensitivity}
\end{figure*}

\subsection{Crop-only training lowers error on the tested cohort}
SupportGain had the lowest clean all-station error among seven selectors in the extended comparison, at 0.801 $\unit$.
That ranking used the same archive and included the cropland test records.
The paired comparison therefore does not include uncertainty from selecting SupportGain.

\begin{figure*}[t]
\captionsetup{width=.96\textwidth}
\centering\includegraphics[trim=120pt 62pt 105pt 0pt,clip,width=.96\textwidth]{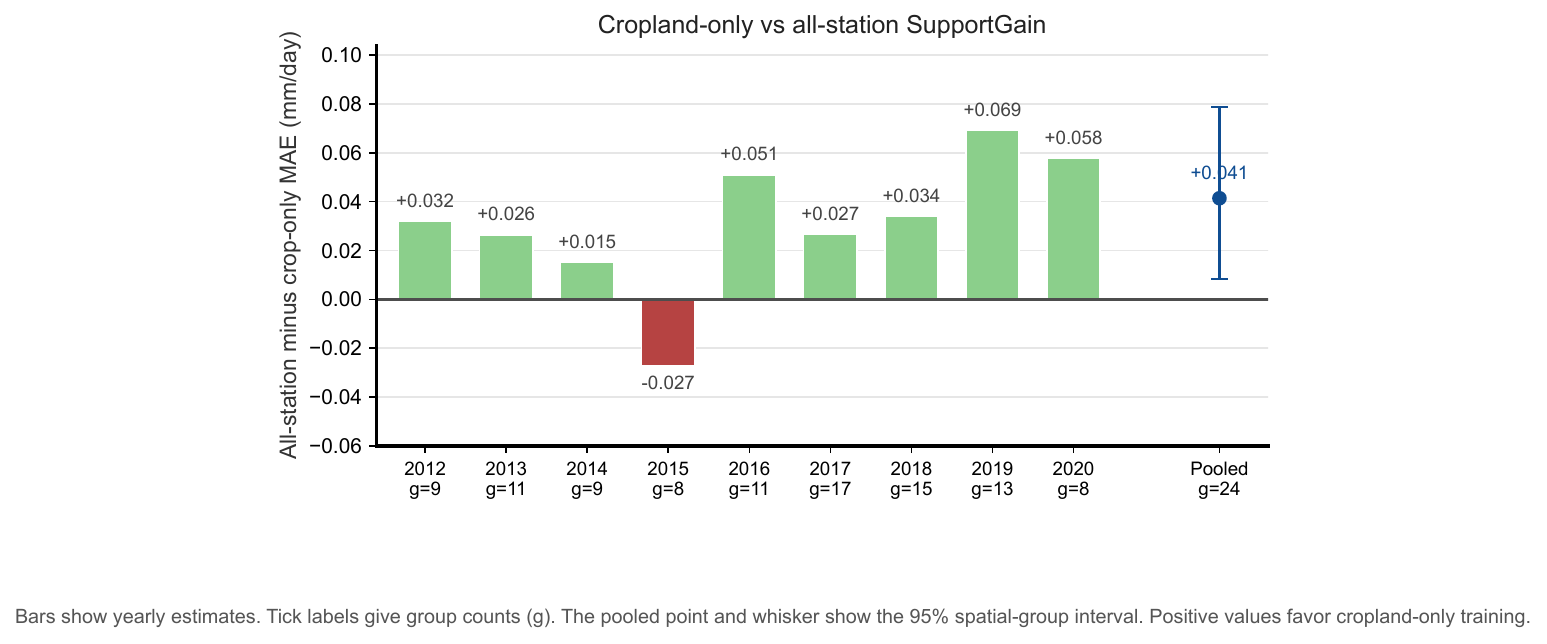}
\par\medskip
\noindent\begin{minipage}{.96\textwidth}
\small\textbf{How to read the result.} Eight of nine yearly estimates favor crop-only training. The years reuse stations and groups, so they show a pattern within this cohort, not independent replications.
\end{minipage}
\caption{Yearly and pooled all-station minus crop-only SupportGain error. Positive values favor crop-only training. The pooled difference is 0.0414 $\unit$ (95\% group-bootstrap interval [0.0085, 0.0788]).}
\label{fig:cropland-training}
\end{figure*}

On the same 3,234 cropland test records from 49 stations and 24 groups, all-station SupportGain scores 0.829 $\unit$ MAE.
Cropland-only SupportGain scores 0.788 $\unit$.
The predeclared difference is 0.0414 $\unit$ (95\% group-bootstrap interval [0.0085, 0.0788]).
The interval uses 2,000 draws and conditions on the fitted models, selectors, and thresholds.
Figure~\ref{fig:cropland-training} shows the annual estimates.

OpenET scores 0.837 $\unit$ on these records.
The post hoc Full comparison also favors crop-only training: 0.777 versus 0.885 $\unit$ MAE.
Its difference is 0.1080 $\unit$ (95\% interval [0.0388, 0.2060]).
The interval uses 2,000 group-bootstrap draws (seed 20261006) across the same 24 groups and conditions on the fitted models.
We reviewed point estimates before this comparison, so it is not independent confirmation or part of the primary family.
Crop-only SupportGain accepts 57.4\% of records; all-station SupportGain accepts 64.1\%.
This result supports population-matched training within this archive, not a general crop or irrigation claim.

Additional risk-coverage curves and exploratory audits remain in the reproducibility package.
\section{Discussion and limits}
The study does not establish a reliable clean-input advantage for support screening.
The 40-test family does not compare SupportGain directly with Gain, and the direct clean differences are small and uncertain.
The fixed weather transformations show how selectors respond to chosen input changes; they do not estimate natural fault rates.

One station shows a severe failure: Gain accepts large corrections on physically invalid weather, while SupportGain falls back to OpenET.
The other 17 groups do not show a broad benefit, and OpenET has the lowest error on those records.
An exploratory audit finds adjusted SupportGain advantages over Gain in 10 of 13 conditions and over MonoGain in nine, but none over AugmentedGain.
The outcomes were visible before that audit, so independent sites with documented faults must test the pattern.
Physical rule violations alone do not prove sensor faults.

Training only on cropland records lowers error on the tested cropland cohort.
However, SupportGain was selected from outcomes on the same archive, including those test records.
The reported interval omits that selection and model refitting.
The result supports a population-matching hypothesis within this archive, not general crop performance.

The rolling-year tests reuse stations and do not measure unseen-site transfer.
Land-cover labels do not identify irrigated fields, and this study does not test irrigation interventions or forecast skill.
All 360 neural fits in the cropland run reach the 120-iteration limit and report convergence warnings.
This limit may affect the fitted predictor and selector.
The measured-weather analysis is exploratory, and the available gridMET data do not provide a device-accuracy scale.
\section{Conclusion}
The study does not identify a generally reliable rule for when a neural correction should change a clean satellite estimate.
It exposes one severe selector failure and finds lower error after crop-only training on the tested cropland cohort, with method-selection uncertainty.
Independent sites and documented input faults must confirm these findings before they support a broader claim.
\section*{Acknowledgements}
I thank Meetpal S. Kukal for research mentorship and critical feedback on an earlier manuscript.
I thank the OpenET and flux-data contributors for making the source datasets available.
Figure styling adapts Chen Liu's \texttt{figures4papers} repository, with attribution and license information in the accompanying materials.
The analysis and conclusions are the responsibility of the author.

\section*{Data, code, and AI assistance}
The source datasets are available through the cited Zenodo records.
The accompanying reproducibility package contains checksums, cohort rows, split assignments, model predictions, protocols, tests, and figure-generation scripts.
The arXiv source package includes the files required to rebuild this manuscript.
The repository contains the model runs, selector protocols, and saved inner and outer predictions.
The numerical environment uses Python 3.13.5, NumPy 2.4.3, pandas 2.3.3, and scikit-learn 1.8.0.
Codex assists with code development, data auditing, analysis, figures, and manuscript drafting.
All reported experimental values come from executed scripts and saved prediction records.

\clearpage
\onecolumn
\pagestyle{fancy}
\appendix
\section{Exploratory measured-weather fault and repair analyses}\label{app:measured-weather}
This appendix reports a corrected three-member analysis on an earlier measured-weather cohort.
It uses seeds 20260713 through 20260715 and recomputes the correction, spread, and support distance for transformed inputs.
We had inspected the original results before correction, so these analyses remain exploratory.
The cohort has 7,758 rows from 84 stations and 62 proximity groups.

\subsection{Weather fault types}
The dose tests multiply wind by 0, 0.447, 1, 2.237, 3.6, 5, or 10.
They add temperature offsets of 0, 5, 10, 20, or 32 degrees Celsius.
Other tests read Fahrenheit values as Celsius and multiply VPD by ten.
Additional wind faults set values to zero, replace missing values with the training median, hold values for seven days, or use a training-derived day-of-year climatology.
The tests keep labels and fitted model parameters fixed.
The source workbook has no instrument accuracy specification, so these runs add no sensor-specification noise.
The ten-member gridMET dose responses appear in Figure~\ref{fig:gridmet-weather-faults}.
Figure~\ref{fig:measured-fault-types} shows the additional measured-weather fault types.

\begin{center}
\includegraphics[width=.88\textwidth]{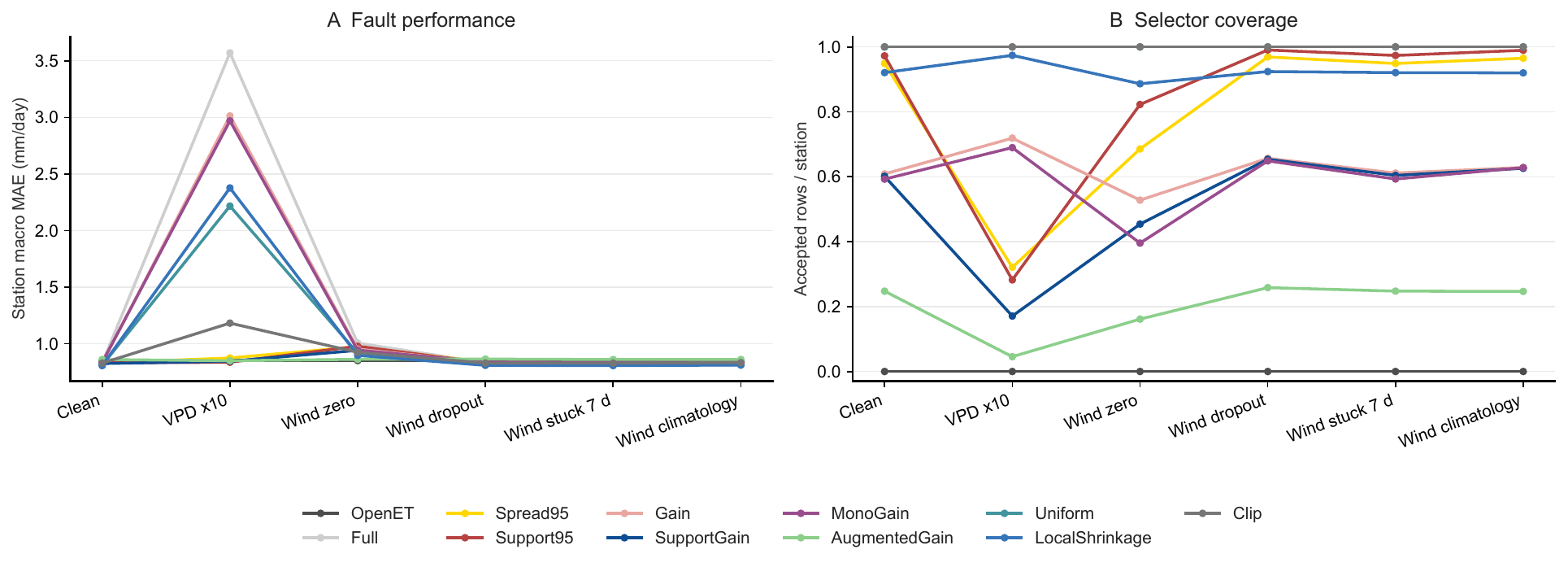}
\captionof{figure}{Exploratory fault-type results from the corrected three-member run. Each transformation keeps fitted model parameters and labels fixed.}
\label{fig:measured-fault-types}
\end{center}

\subsection{Selector repairs}
Three controls test alternative correction choices.
\emph{Uniform} selects a single multiplier from $\{0,0.25,0.5,0.75,1\}$ by inner station MAE.
\emph{LocalShrinkage} estimates the conditional moments $\E[(y-o)g\mid x]$ and $\E[g^2\mid x]$ with two gradient-boosted regressors.
It multiplies $g(x)$ by $\lambda(x)=\operatorname{clip}(\widehat{\E}[(y-o)g\mid x]/\widehat{\E}[g^2\mid x],0,1)$, or sets $\lambda(x)=0$ when the denominator is zero.
The ratio targets squared error, not the primary absolute-error score.
The regressors use inner out-of-group predictions, equal station weights, and the Gain feature set.
\emph{Clip} limits weather inputs to the outer-training 1st and 99th percentiles at inference.

LocalShrinkage has clean station-macro MAE 0.806 mm/day; SupportGain has 0.825.
The paired interval for its improvement is [0.001, 0.037] across 62 groups.
The Holm-adjusted p-value is 0.154 across eleven post hoc comparisons.
LocalShrinkage and Uniform differ by 0.003 mm/day, with an interval of [-0.009, 0.019].

Under tenfold wind, MonoGain has station-macro MAE 1.230 mm/day and Gain has 1.242.
SupportGain has 0.853 on these same probe rows.
Corrected AugmentedGain has MAE 0.860 on clean rows, against 0.825 for SupportGain.
Under tenfold wind, AugmentedGain has 0.862 and SupportGain has 0.853.
The 95\% interval for the AugmentedGain improvement is [-0.0231, 0.0037] mm/day.
None of its four comparisons with SupportGain passes Holm correction.
These results do not identify a confirmed selector repair.

\subsection{Spatial holdout sensitivity}\label{app:spatial-holdout}
This post hoc analysis pools five spatial folds on the clean gridMET cohort.
It uses 16,366 rows from 151 stations and 102 spatial groups.
Each fold's fit excludes every row from its test groups.
Each row receives one prediction from a model that excludes its station.
Training uses all dates from other groups, so this does not test forecasting.
The selector and condition choices were visible before this analysis.

\begin{center}
\begin{minipage}{.96\textwidth}
\centering
\small
\setlength{\tabcolsep}{3pt}
\captionsetup{width=\linewidth}
\captionof{table}{Gain-minus-SupportGain station-macro MAE (mm/day) across five spatial folds. Positive values favor SupportGain. Intervals are simultaneous across the five test conditions.}
\label{tab:spatial-holdout}
\begin{tabularx}{\linewidth}{@{}Xrr@{}}
\toprule
Test input & Difference & Simultaneous 95\% interval \\
\midrule
Clean & 0.0059 & [-0.0069, 0.0186] \\
Wind x2.237 & 0.0384 & [0.0015, 0.0752] \\
Wind x3.6 & 0.1478 & [0.0697, 0.2260] \\
Fahrenheit as Celsius & 0.4899 & [0.3454, 0.6344] \\
VPD x10 & 1.3698 & [0.4886, 2.2511] \\
\bottomrule
\end{tabularx}
\end{minipage}
\end{center}

The group bootstrap uses 2,000 draws with seed 20261060.
No p-values were calculated.
These intervals condition on fitted models and omit retraining uncertainty.
The four unit-error conditions preserve labels and model parameters.
They do not estimate natural fault prevalence.

\begin{figure*}[t]
\centering
\includegraphics[width=.94\textwidth]{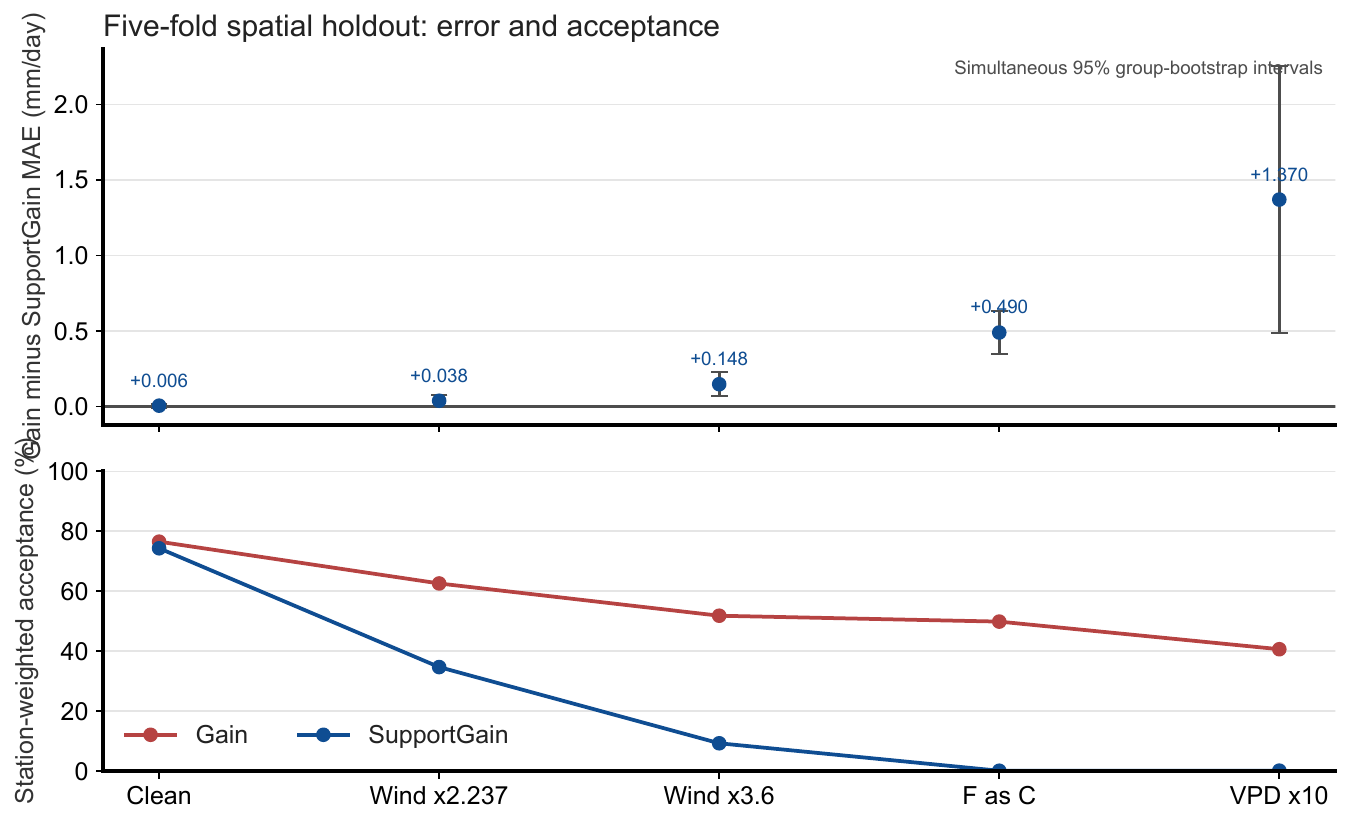}
\caption{Gain-minus-SupportGain MAE difference and acceptance on held-out spatial groups. Positive values favor SupportGain. The upper panel shows simultaneous 95\% group-bootstrap intervals.}
\label{fig:spatial-holdout}
\end{figure*}

The full prediction tables, risk-coverage curves, matched-budget analyses, original benchmark, and earlier covariate ablations remain in the reproducibility package.
\clearpage\twocolumn
\pagestyle{fancy}
\balance
\begingroup\small
\bibliographystyle{plainnat}
\bibliography{references}
\endgroup
\end{document}